\documentclass[letterpaper, 10 pt, conference]{ieeeconf}  
\usepackage{graphicx}
\usepackage{tabularx}
\usepackage{multirow}
\usepackage{xcolor}
\usepackage{amsmath,amssymb}
\usepackage{float}
\usepackage{booktabs}
\usepackage{hyperref}

\IEEEoverridecommandlockouts                              

\title{\LARGE \bf 
Implicit Neural Representations for Breathing-compensated Volume Reconstruction in Robotic Ultrasound 
} 

\author{Yordanka Velikova$^{1 *}$, Mohammad Farid Azampour$^{1,3,5 *}$, Walter Simson$^{2}$, Marco Esposito$^{4}$, and Nassir Navab$^{1}$
    \thanks{*Shared contribution.}%
    \thanks{$^{1}$    Computer Aided Medical Procedures, Technical University of Munich, Munich, Germany
        {\tt\small email: dani.velikova@tum.de}}%
    \thanks{$^{2}$ Department of Radiology, Stanford University School of Medicine, Stanford, CA, USA}%
    \thanks{$^{3}$  Department of Electrical Engineering, Sharif University of Technology, Tehran, Iran}
    \thanks{$^{4}$  ImFusion GmbH, Munich, Germany}
    \thanks{$^{5}$ Munich Center for Machine Learning}
    \thanks{Github Repository: \url{https://github.com/danivelikova/INR-3DUS}}
}

\begin{document}

\maketitle
\thispagestyle{empty}
\pagestyle{empty}

\begin{abstract}
Ultrasound (US) imaging is widely used in diagnosing and staging abdominal diseases due to its lack of non-ionizing radiation and prevalent availability. However, significant inter-operator variability and inconsistent image acquisition hinder the widespread adoption of extensive screening programs.
Robotic ultrasound systems have emerged as a promising solution, offering standardized acquisition protocols and the possibility of automated acquisition. Additionally, these systems enable access to 3D data via robotic tracking, enhancing volumetric reconstruction for improved ultrasound interpretation and precise disease diagnosis.

However, the interpretability of 3D US reconstruction of abdominal images can be affected by the patient's breathing motion. This study introduces a method to compensate for breathing motion in 3D US compounding by leveraging implicit neural representations. Our approach employs a robotic ultrasound system for automated screenings. To demonstrate the method's effectiveness, we evaluate our proposed method for the diagnosis and monitoring of abdominal aorta aneurysms as a representative use case.

Our experiments demonstrate that our proposed pipeline facilitates robust automated robotic acquisition, mitigating artifacts from breathing motion, and yields smoother 3D reconstructions for enhanced screening and medical diagnosis.

\end{abstract}

\section{Introduction}

Medical ultrasound (US) is widely employed in clinical settings to diagnose conditions of internal tissues and organs. Its real-time imaging offers clinicians instant feedback. In comparison to other imaging methods like X-ray or CT, US presents several advantages: it's radiation-free, portable, and cost-effective \cite{gibbs2011ultrasound}. This, combined with its high soft-tissue contrast, ensures both safety and efficiency in patient care \cite{szabo2004diagnostic}.
Ultrasound has become an essential tool for screening and monitoring of the abdominal region \cite{usdiag, usaortameas}. In particular, for individuals aged 50 and above, routine US screening of aorta is recommended, as they might be at risk of developing aneurysms while being asymptomatic. Abdominal Aortic Aneurysm (AAA) is a condition marked by an enlargement in the aorta, which can lead to serious health implications if left untreated \cite{ullery_epidemiology_2018}. If an enlargement is detected, the frequency of follow-up screens increases based on its growth rate, ensuring timely interventions \cite{hartshorne2011ultrasound}.
Consistent ultrasound screenings have shown a significant reduction of premature death from AAA in men aged 65 and older~\cite{Guidelinesaaa}. The primary goal of these screenings is to accurately measure and assess the aneurysm's size and dimensions. 

However, the repeatability of image acquisitions can vary largely between clinicians \cite{li_overview_2021} as image quality is strongly based on clinician experience and factors such as probe placement and contact during scanning can reduce inter-acquisition consistency. 
\begin{figure}[t!]
    \centering
    \includegraphics[width=0.9\columnwidth]{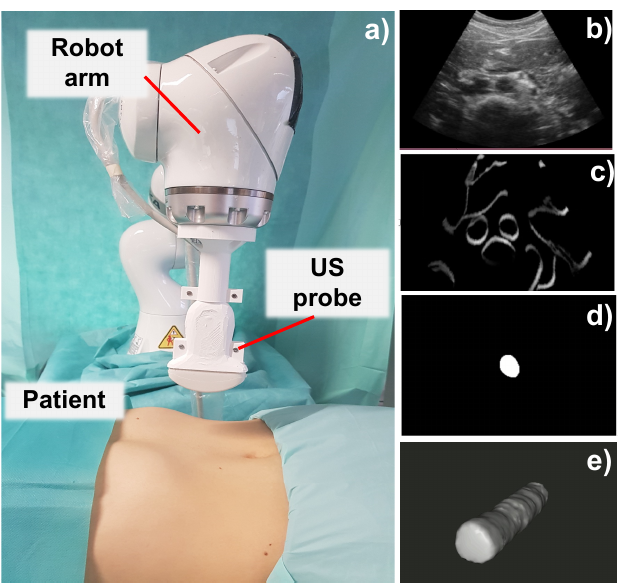}
    \caption{Overview of the proposed system: a) we use a robotic arm manipulator with a convex ultrasound probe attached to the end-effector to scan the patient, b) Live B-mode, c) Intermediate image, d) Aorta segmentation, e) 3D Reconstruction.}
    \label{Fig:overview_setup_patient}
\end{figure}
\begin{figure*}[th]
    \centering
    \includegraphics[width=0.9\textwidth]{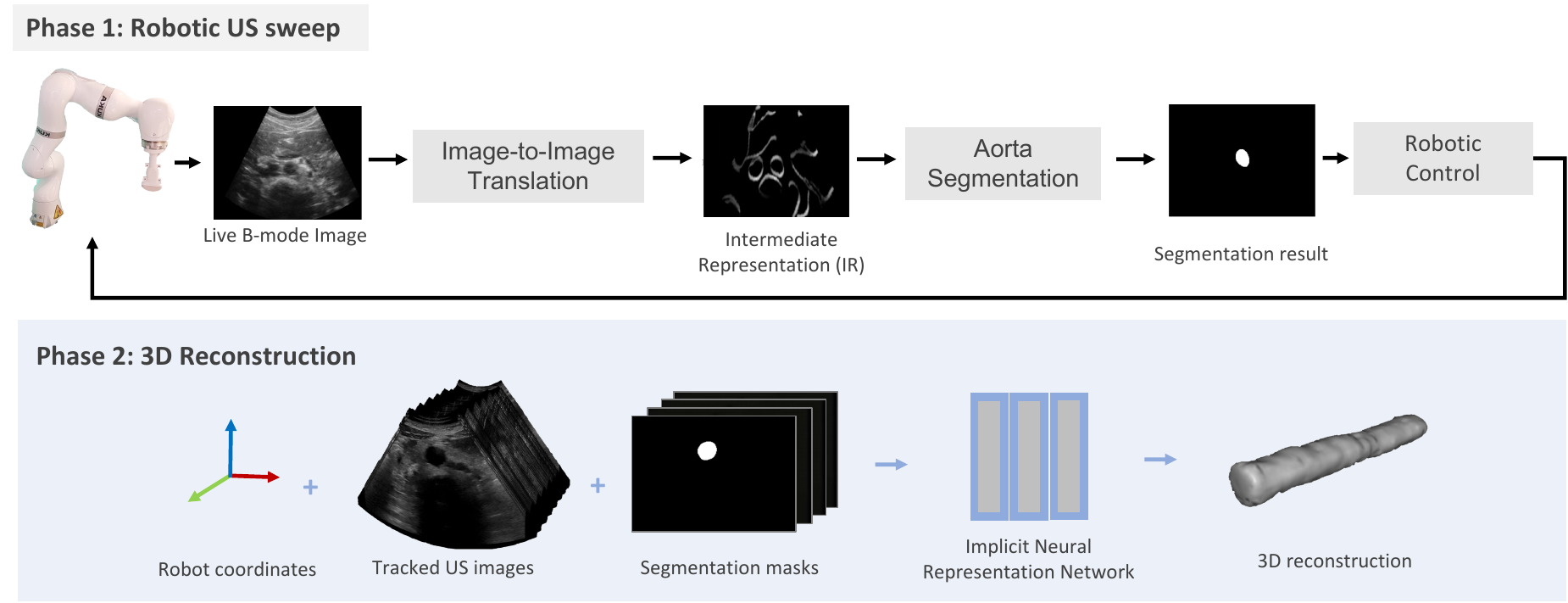}
    \caption{Overview of the pipeline. Phase 1: Robotic ultrasound acquisition with real-time image segmentation; the robot's trajectory is adjusted based on the segmentation. Phase 2: The acquired sweep and segmentation train the INR model, which is sampled to produce a dense aorta point cloud. Post-processing of this cloud yields the final mesh.}
    \label{Fig:overview_method}
\end{figure*}
These constraints highlight the necessity to automate US acquisitions. Recent works have exploited the use of robotic systems for manipulating the US probe autonomously, thus providing standardized imaging results and reducing operator dependency~\cite{chatelain,hennespr}.
Solutions for automating US acquisitions have been a focus of study for more than two decades, ranging from the creation of customized end-effectors for guiding US probes to the introduction of complete imaging and navigation systems~\cite{JiangMedicalIA}.

Recent work demonstrates an automated workflow for screening of tubular structures using a robotic US system \cite{Jiang2020AutonomousRS}. It is based on real-time image feedback and employs active perception to control robotic motion. They estimate the centerline and radius of a tubular structure and perform robotic control in real-time in response to the incoming US images. In addition, they extract the 3D point cloud of the segmented tube and calculate its centerline. However, the method lacks generalizability on in-vivo images.

Furthermore, employing robotic ultrasound enables 3D reconstruction of the aorta. Access to three-dimensional representation can help clinicians compare the progression of aneurysms.
Without explicit access to 3D information, physicians need to mentally reconstruct the 3D anatomy, which is a demanding and challenging task, especially for less experienced ones. Having a 3D reconstruction of the aorta volume can give more insights to clinicians and help them to evaluate the diameter of the aneurysm not only in the axial 2D plane but also in any arbitrary tilted plane. 

However, creating a consistent US compounding faces hurdles like patient breathing motion. Virga et al. \cite{forcecompliantRUS} proposed solutions to automate trajectory planning and address motion from breathing. Despite the strides, motion artifacts remain unresolved. In robotics, breathing compensation remains a prominent topic, with gating control techniques and user-in-the-loop systems ensuring stable acquisitions \cite{ecg, breatg}.

Lately, the rise of implicit neural representation (INR) stands out, especially for rendering continuous 3D volume representations. In medical imaging, INRs encapsulate anatomical data compactly while preserving resolution. Typically, a Multilayer Perceptron (MLP) in INRs infers pixel or voxel intensity based on their position.
Adding to the advancements in this domain, Sitzmann et al.'s SIREN~\cite{sitzmann2020implicit}, which uses periodic functions over traditional activations, has achieved impressive results in capturing scene nuances.
Khan et al.~\cite{khan2022implicit} utilized SIREN to obtain a segmentation mask that converges quickly and, by definition, supports super-resolution. Gu et al.~\cite{gu2022representing} employed SIREN for US imaging by integrating the SIREN-based representation with meta-learning. This approach facilitates learning an INR for a new patient using a set of sparse inputs. Continuing the application of SIREN in US imaging, Song et al.~\cite{song2022development} aimed to enhance the segmentation of the carotid artery in US images. They hypothesized that given an US sweep and the segmentation output from a trained 2D U-net, the MLP could infer both pixel intensity and label. Their findings suggest that an INR trained in this manner can produce smoother label maps, preserving the curvature of the carotid artery, even with tracking data errors. In our work, we also delve into INR's capabilities to enhance our robotic US acquisition and segmentation. In a recent work, exploiting the power of INRs combined with a physics-based rendering module, Wysocki et al.~\cite{wysocki2024ultra} showed how Ultra-NeRF can generate highly accurate B-mode ultrasound images through learning US-specific parameter maps which account for view-dependent variations inherent to ultrasound imaging.

In this work, we present a robotic US system for autonomous aortic scan with breathing compensation for improved 3D US volume reconstruction. The robot control strategy does not require pre-operative registration with other imaging modalities in order to perform the navigation, nor external tracking systems or cameras, but only a single manual initialization step by the medical personnel. Once the probe is placed such that the aorta is visible in the US image, it can follow it to cover the whole region of interest without further intervention. By leveraging a recent method for aorta segmentation incorporating the concept of intermediate representation \cite{CACTUSS}, it can be directly applied to real US images, while being trained only on simulated ones. 

Our key contribution is the development of a breathing-compensated US reconstruction method. This technique harnesses an INR model, trained using the gathered tracked US images and their respective segmentation masks. This network not only crafts a patient-tailored continuous 3D function from a subset of the images but also compensates for noisy or missing segmentation frames to generate a smooth, finely detailed 3D aorta reconstruction. 

\section{Methodology}

The proposed robotic ultrasound screening system is composed of a robotic manipulator and a convex US probe rigidly attached to the robot's end-effector, see Fig \ref{Fig:overview_setup_patient}. 
The overall workflow setup of the method consists of two main steps: robotic ultrasound sweep and 3D reconstruction, as shown in Fig. \ref{Fig:overview_method}, which are explained in detail below.

\subsection{Image acquisition and segmentation}\label{sec:imaging}
Prior to beginning the acquisition procedure, the probe must be manually placed on top of the patient's abdomen, usually in the upper part, where the aorta is visible in the US image. Our navigation during the acquisition depends on robust real-time segmentation of the aorta.
To achieve this, we utilize a recently proposed pipeline called CACTUSS~\cite{CACTUSS}, which is tailored for the task of aorta segmentation in US images. CACTUSS proposes a novel concept of IR between CT and US, and it is trained on a large set of simulated intermediate images from CT labelmaps. IR space is introduced to transfer both simulated and real images to this space by using a domain adaptation network. Then, a segmentation network is trained on the transferred simulated images in the IR space.
Training in the IR space allows for direct segmentation of real US images, even though the network has been trained only on simulated ones, and it shows great generalization capabilities across patients.

As depicted in Fig.~\ref{Fig:overview_method} each US B-mode image is first passed through the pre-trained image-to-image translation network, which translates it to the IR image space. The translated image is then passed to the second network, which gives the segmentation result. Finally, the prediction output is used to control the robot's movement in real-time, enabling visual servoing and autonomous US acquisition. 

\subsection{Navigation pipeline}

Along the anteroposterior axis, the position of the end-effector is controlled using a constant force, ensuring that the probe maintains a consistent connection to the skin for optimal image quality. For longitudinal and transverse axes, a visual-based control pipeline is used to update the position of the robot's end-effector. 
During the scan, a sequence of two-dimensional B-mode frames is collected from the US machine, and the pose is updated in real-time based on the output of the segmentation algorithm.

\begin{figure}[th!]
    \centering
    \includegraphics[width=0.6\columnwidth]{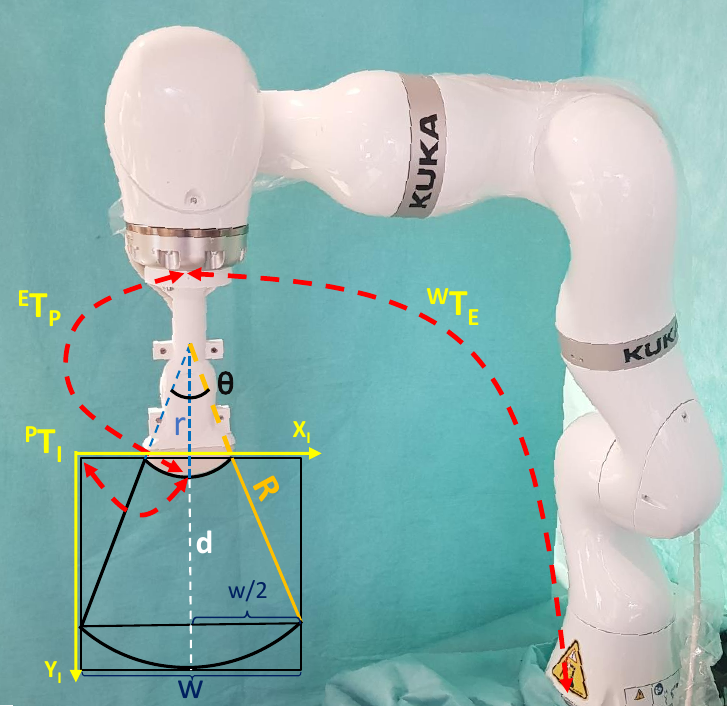}
    \caption{Robot transformations.}
    \label{Fig:transformations}
\end{figure}

To avoid navigating to a wrong position as a result of incorrect segmentation output, we ensure that the aorta is always maintained in the center of the image. We keep track of the center of the previous segmentations, and the robot moves along the transverse axis to ensure that the aorta is always centered.   
Consequently, a step on the longitudinal axis is performed, and the robot moves until no aorta is visible, allowing a complete sweep to be obtained.
To enable our control pipeline, it is necessary that the images are transformed into the coordinate frame of the robot. For every image frame, the full chain of transformations from robot base to image frame is calculated (see Fig. \ref{Fig:transformations}):
\begin{equation}
    {^W}T_{I} = {^W}T_{E} \cdot {^E}T_{P} \cdot {^P}T_{I}
\end{equation}
where ${}^PT_{I}$ maps from the origin of the image to the probe by converting image pixels to millimeters, ${^E}T_{P}$ is the transformation from the tip of the probe to the end effector, which is known from the 3D model of the mount, ${^W}T_{E}$ is the transformation from the robot's end-effector to the robot base and is calculated using the robot's kinematic model, which is given by the manufacturer.

\begin{figure*}[h!]
    \centering
    \includegraphics[width=0.8\textwidth]{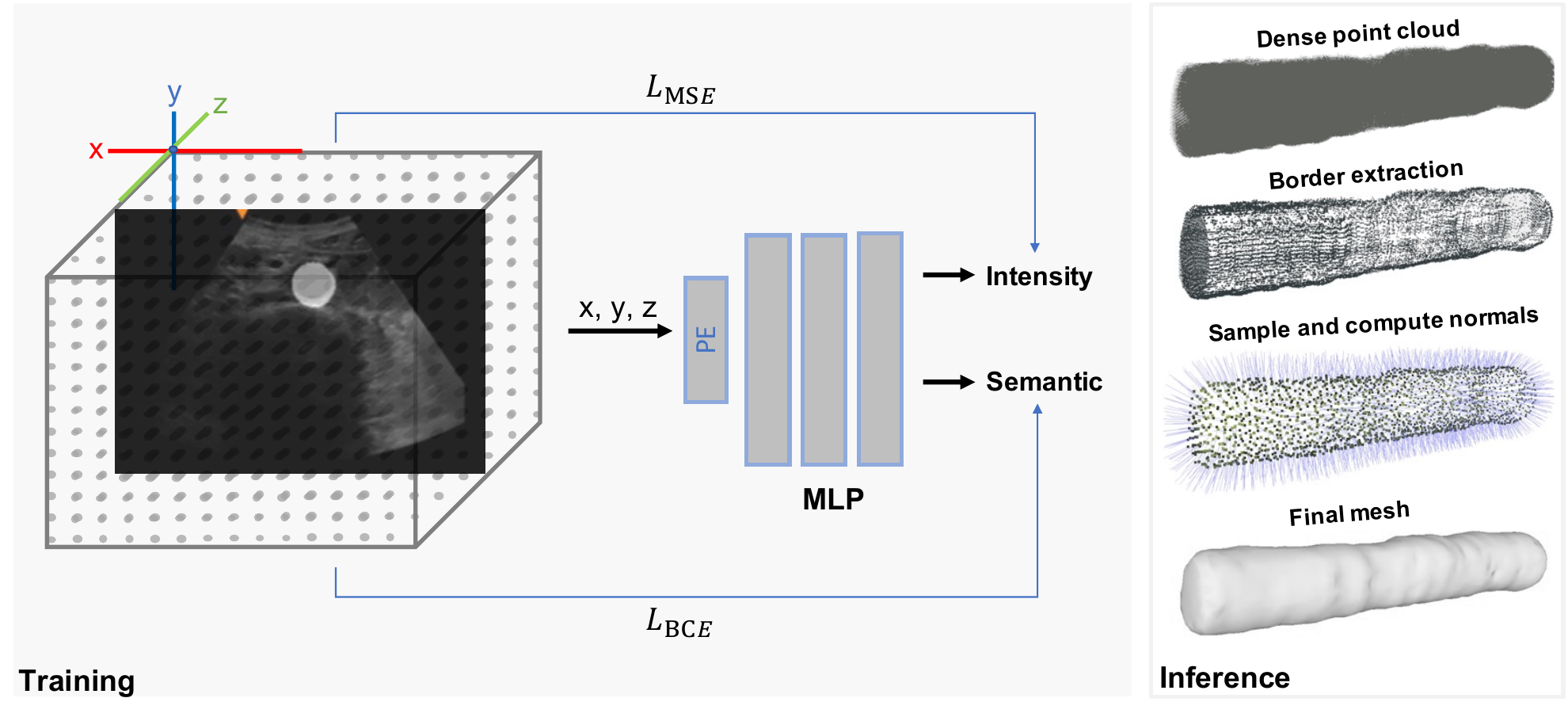}
    \caption{Schematic of the INR model. The model accepts a normalized 3D voxel position (ranging from $-1$ to $1$) as input and produces two outputs: the intensity and the semantic label of the voxel. For training, we leverage the ultrasound (US) sweep alongside its corresponding segmentation. During inference, we sample across all points, yielding a dense point cloud from the semantic results. This is subsequently post-processed to generate the final mesh.}
    \label{Fig:inr_archit}
\end{figure*}

\textbf{Calibration}\label{methodology:calibration}
To map between pixels and millimeters, we use the dimensions of the image in pixels ($W \times H$) and the opening angle of the probe $\theta$. The imaging depth $d$ in mm is typically specified by the user only once in the beginning. The mapping can be calculated by the matrix ${^W}T_{I}$:
\begin{equation}
    {^W}T_{I} = 
    \begin{bmatrix}
        \frac{w}{W} & 0 & 0 & -\frac{w}{2}\\
        0 & 0 & -1 & 0 \\
        0 & \frac{d}{H} & 0 & -y \\
        0 & 0 & 0 & 1 \\
    \end{bmatrix}
\end{equation}
where  $W$ and $H$ are the width and height of the image in pixels, $w$ is the corresponding offset in mm along the width of the image, derived by $w = 2 R \sin(\theta /2)$, where $R = d+r$ is the radius defined as the sum of the depth of the image $d$ and the radius $r$ from the origin of the probe to its tip.
Additionally, a small offset $h$ along the height of the image is calculated from the tip of the probe to the origin of the image: $  h = r - \cos(\theta /2)r$.

\subsection{3D Volume Reconstruction via the INR Model} 

To utilize INR-based models for volume reconstruction, we first acquire a set of observations from the 3D space, along with their corresponding poses. In our study, the observation refers to the US sweep, while the pose is derived from the robotic tracking stream. The aim of our INR model is not just to encode the anatomical intensity values but also the semantic label of the aorta. Analogous to our observation in the intensity domain, we utilize the segmentation results obtained during the data acquisition phase (refer to section~\ref{sec:imaging}) as our observations from the semantic domain. 
Similar to~\cite{song2022development}, our INR model has been adapted to produce both volume intensity and semantic attributes. Through this approach, we ensure the learning of a continuous smooth function in the semantic domain, which solidifies attributes vital for vessel segmentation.
The INR model produces the output:
\begin{equation}
    {\bf y} = (s, i) \in \mathcal{Y} \subseteq \mathbb{R}^{M+1}
\end{equation}
where \( s \in \mathcal{S} = \mathbb{R}^M \) represents the semantic vector and \( M \) denotes the number of semantic classes. \( i \in \mathbb{R} \) stands for the volume intensity. An INR model is trained to infer {\bf y} for each given position in the 3D domain. In particular, if we define a location vector as:
\begin{equation}
    {\bf x} = (x, y, z) \in \mathcal{X} \subseteq \mathbb{R}^3,
\end{equation}
The trained INR is function that maps {\bf x} to our desired output:
\begin{equation}
    {\bf y} = f_{\Theta}({\bf x})
\end{equation}
Here, \( f_{\Theta} \) designates the overall INR model, with \( \Theta \) representing its trained weights. When trained, \( f_{\Theta} \) encodes the entire intensity and semantic information as a continuous function that can be sampled at any resolution. The design of our network can be found in Fig.~\ref{Fig:inr_archit}.

To preserve the high-frequency information of the volume intensity, we apply Position Encoding (PE)~\cite{sitzmann2020implicit}, which maps each 3D coordinate into a more expansive dimensional space. The encoding of \( {\bf x} \) can be achieved with the function \( \phi: \mathbb{R} \rightarrow \mathbb{R}^{2L} \):
\begin{align}
    \gamma(p) = &( sin(2^{0}\pi p), cos(2^{0}\pi p), \ldots, 
    \\ & sin(2^{L-1}\pi p), cos(2^{L-1}\pi p)) \nonumber
\end{align}
Where \( p \) refers to the normalized values (ranging from $-1$ to $1$) of the 3D positions \( (x, y, z) \) respectively.

The main component of \( f_{\Theta} \) is an MLP. In this MLP, the path for achieving the two desired outputs from \( f_{\Theta} \), meaning the intensity and the semantic information are shared except for the last linear layer. As in~\cite{sitzmann2020implicit}, to depict the high-frequency domain, the SIREN activation function is employed.
Further, we define the volume intensity loss as \( \mathcal{L}_i \) and the semantic loss as \( \mathcal{L}_s \): 

\begin{equation}
    \mathcal{L}_i = \sum_{x \in \mathcal{X}} ||\hat{i}(x) - i(x)||^2
\end{equation}
\begin{equation}
    \mathcal{L}_s = - \sum_{x \in \mathcal{X}} \left[ \sum_{m=1}^{M} s^m(x) \log \hat{s}^{m}(x) \right].
\end{equation}

In the above, \( \mathcal{X} \) represents the input location space, and \( \hat{i}(x) \) and \( i(x) \) denote the predicted and ground truth volume intensities, respectively. Similarly, \( \hat{s}^{l}(x) \) and \( s^l(x) \) indicate the semantic probability for class \( l \) in the predictions and ground truth, respectively. Combining these losses, the overall training loss can be defined as:

\begin{equation}
    \mathcal{L} = \mathcal{L}_i + \mathcal{L}_s
\end{equation}

During training, our semantic observations are derived from the CACTUSS segmentation pipeline. However, due to breathing motions and inaccuracies in translating to the IR space, some slices might contain erroneous segmentations. Including these in training could compromise the accuracy of our INR model. To mitigate this, for each slice, we retain only the largest connected component, setting all else to zero. Moreover, by monitoring the aorta's radius across previous slices and employing a moving average approach, we obtain an estimated radius for the current slice. If there's a significant discrepancy between this estimate and the actual value, we label the slice as incorrectly segmented.

For each slice determined to be correctly labeled, both semantic and intensity losses are calculated for its respective voxels. For those identified as mislabeled, only the intensity loss is computed. This strategy harnesses valuable geometric information from correct slices while sidestepping potential pitfalls posed by erroneous semantic labels.

Once the INR network is trained, inference is performed as illustrated in Fig.~\ref{Fig:inr_archit} to achieve the final 3D volume reconstruction. We begin by sampling new points to form a dense point cloud from our predictions. We then identify boundary points by fitting a convex hull to each slice. Using the furthest point sampling method, points are sampled, and their normals are computed. Lastly, we employ the Poisson surface reconstruction~\cite{kazhdan2006poisson} to produce the final mesh.

\section{Setup}
\subsubsection{Hardware}

The system consists of a curvilinear transducer 5C1 ACUSON Juniper US machine (Siemens
Healthineers, Germany) which is attached to the end-effector of a robotic manipulator (KUKA LBR iiwa 14 R820, KUKA Roboter GmbH, Germany) using a 3D-printed  probe holder. We maintain a constant low force (5N) and stiffness(200 N/m) to allow comfort for the patient during breathing. The robot is controlled via iiwa stack and ROS. 
To access the images, a frame grabber (Epiphan Video, Canada) is used and visualized in real-time on ImFusion Suite (ImFusion GmbH, Germany). The image depth is set to 10cm. For working with the data, ethics approval was obtained.\footnote{Approval was obtained from the ethics committee of the Technical University of Munich.}

\subsubsection{Training Details}
The INR network consists of an MLP with 6 layers, each of size 256, and employs positional encoding of length 10. The network is trained for each dataset individually. 
The batch size, in our case, is the number of image slices and is set to 1. We trained for 1000 epochs with a learning rate of 1e-3 and Adam optimizer.

\section{Evaluation}

To achieve an optimal 3D reconstruction free from breathing-induced distortions, we take into account images captured during the exhale phase only. Breathing movements can introduce errors in segmentation masks and generate noise. The stability of images during exhale allows for more accurate segmentation prediction. We conducted two primary experiments: breath-hold and normal breathing, which are detailed in the subsequent sections. In both experiments, we compared our novel INR pipeline with conventional techniques. 
The breath-hold experiment, being user-controlled to capture images only during the exhale phase, acts as a reference point for our model evaluation. Fig.~\ref{Fig:button_vs_breathing} illustrates both modes. Additionally, we compute the Laplacian and report the overall roughness of the final 3D volume surfaces.

\subsection{Breath-hold with button}
In this experiment, a volunteer is instructed to press and hold a button after a deep exhale. As this action is performed, the robot moves, and the US images and segmentation masks are stacked with their 3D location based on the robotic tracking data. In Fig.~\ref{Fig:button_vs_breathing}, the left side shows the trajectory of the images in 3D, where a straight trajectory is observed during the button hold. Upon releasing the button, the robot halts, allowing the volunteer to breathe freely. This pause is marked by the peaks in the trajectory, and no images are captured during this time.

We train our INR model using the acquired 2D B-modes, their corresponding segmentations, and 3D coordinates. For this experiment, every third image from the scan is used to build the training set. We compare the results of our model to a conventional linear interpolation method, implemented in ImFusion Suite\footnote{ImFusion GmbH, Munich, Germany}.

\subsection{Free breathing}

\begin{figure}
    \centering
    \includegraphics[width=0.99\columnwidth]{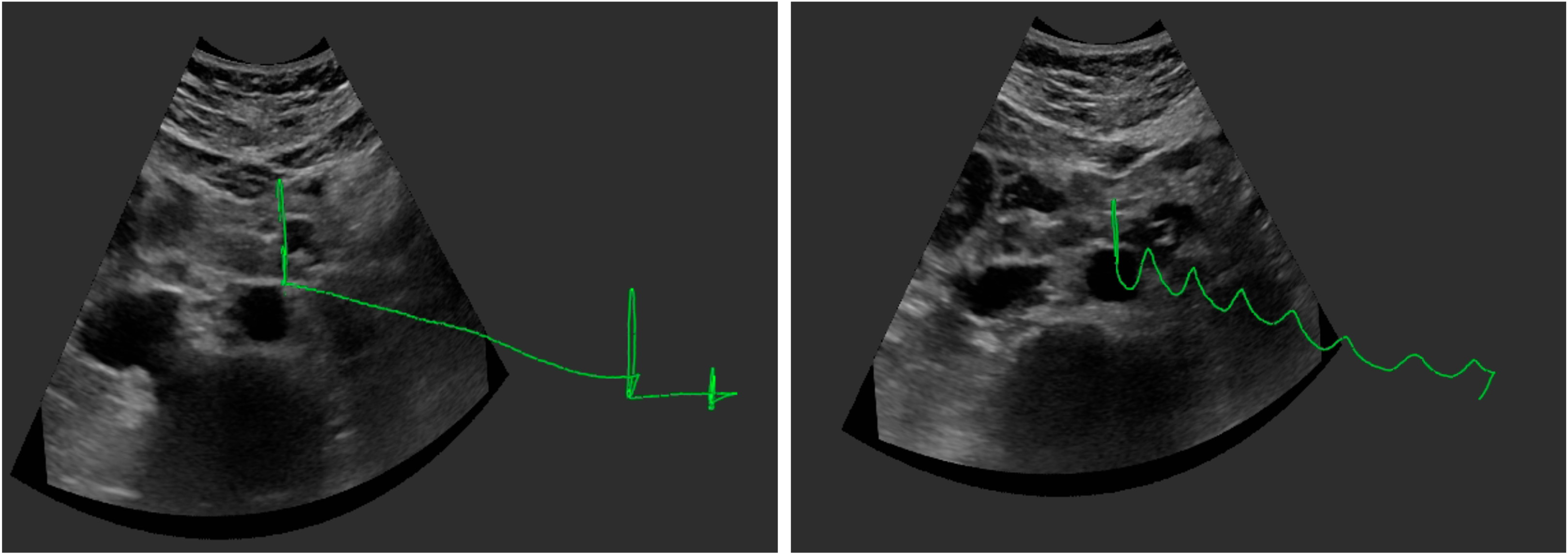}
    \caption{Left:breath-hold mode, right: free breathing mode.}
    \label{Fig:button_vs_breathing}
\end{figure}

The second experiment is performed during free breathing, without any restrictions on the user. The right side of Fig.~\ref{Fig:button_vs_breathing} displays the breathing trajectory. Once the sweep is completed, the segmentations are again compiled in 3D. The top of Fig.~\ref{Fig:breathing_mode} portrays the 3D volume right after the acquisition without any breathing adjustments, showing distortions due to respiratory movements.
To derive a high-quality 3D reconstruction and compare it with the breath-hold experiment, we subsequently extract images only from the exhale phase(red). Using the 3D trajectory from the robot, we identify all local minima, selecting only the images within a narrow range surrounding each minima. These images are used as training data for the INR network. The results from the network are again compared with the conventional linear interpolation technique.

\subsection{Second volunteer}
We demonstrate the generalizability of our algorithm using a US sweep acquired from a second volunteer. 
It's important to note that INR-based methods are trained specifically for each patient, necessitating retraining from scratch for every new individual. Since the INR is patient-specific, the generalization of our full pipeline depends on the capabilities of the underlying segmentation method. Our segmentation method (CACTUSS) has been tested on multiple subjects, and, as demonstrated in the original paper, shown to be generalized to unseen patients. By testing on a second volunteer, we confirm that while the INR network needs individualized training, the full pipeline itself is broadly applicable. The qualitative results for this volunteer, under both breath-hold and free-breathing scenarios, are presented in Fig. \ref{Fig:button_mode} and \ref{Fig:breathing_mode} respectively, and quantitative results are reported in Table \ref{Tab: Results Phantom Detection}.

\section{Results and Discussion}

\begin{figure}
    \centering
    \includegraphics[width=\columnwidth]{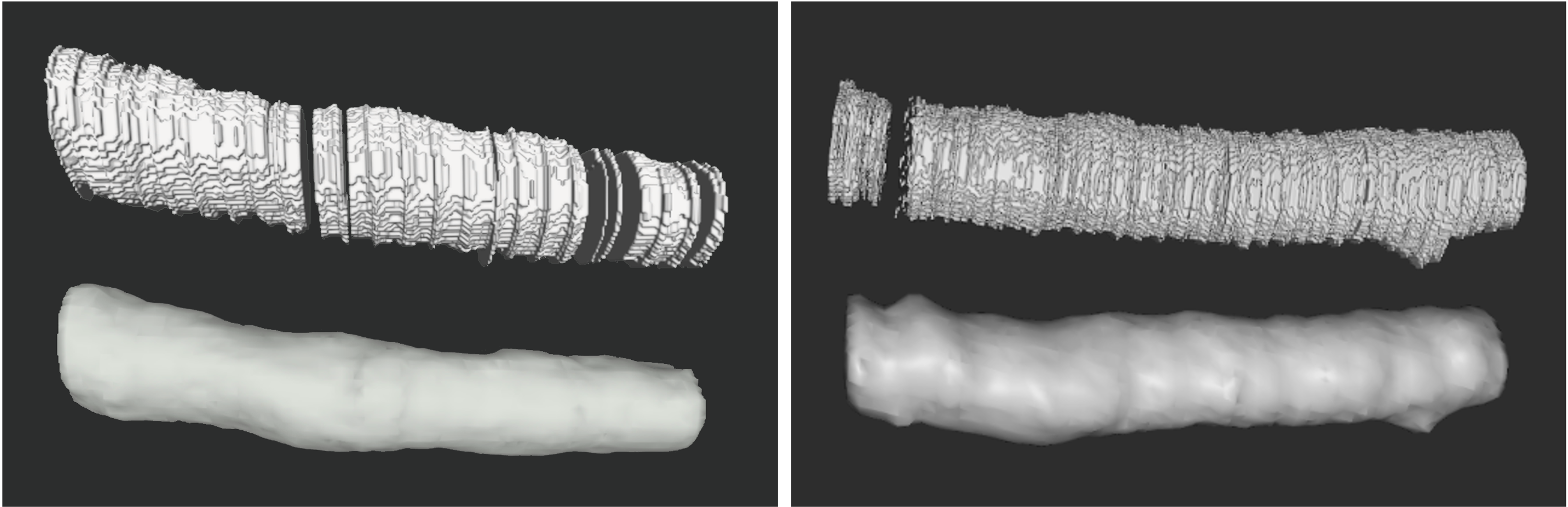}
    \caption{3D volumes from two volunteers from conventional (top) and INR method (bottom) from the Breath-Hold Button experiment. Conventional methods fail in interpolating larger gaps, while our INR-based approach offers smoother interpolation between frames.}
    \label{Fig:button_mode}
\end{figure}

Fig.~\ref{Fig:button_mode} illustrates the 3D volumes from the breath-hold experiment on both volunteers using both conventional and INR methods. Our technique provides a notably smoother 3D reconstruction than traditional approaches, which often fails when interpolating larger gaps. By learning a patient-specific continuous function, our method smoothly interpolates between frames, resulting in an enhanced 3D reconstruction. 

\begin{figure}
    \centering
    \includegraphics[width=\columnwidth]{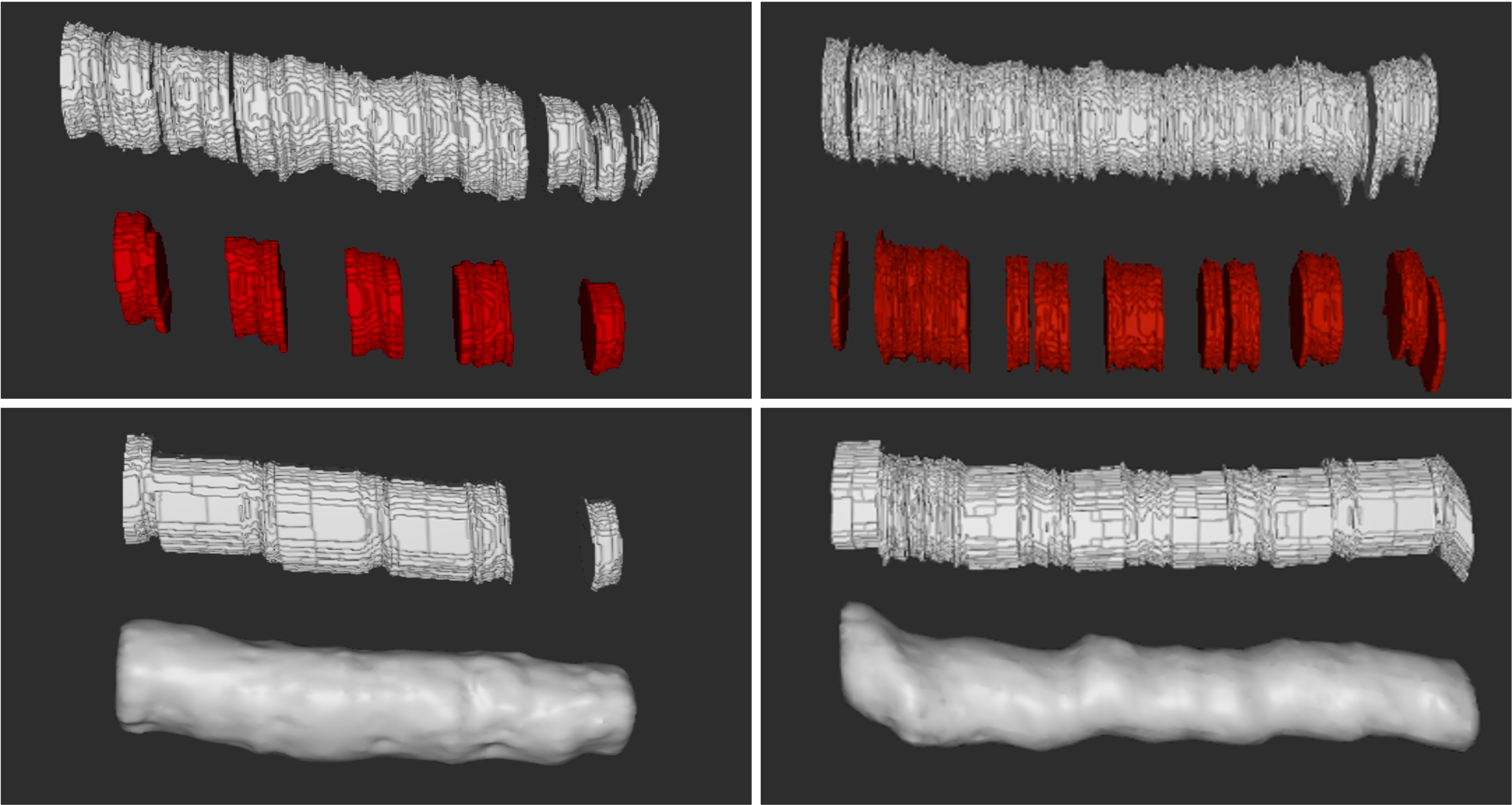}
    \caption{Top image: the 3D volumes acquired after the free breathing experiment from two volunteers, showing distortions due to respiratory movements. Red: images only from the exhale phase, used for training of the INR. Bottom image: 3D volume generated with conventional method, 3D volume reconstruction result from our network.  
    }
    \label{Fig:breathing_mode}
\end{figure}

\begin{table}
\centering
\caption{Comparison of mesh smoothness under two breathing modes: "Breath-Hold Button" and "Free Breathing". Results indicate that our pipeline produces smooth outputs for both modes. Further, utilizing the INR with filtered input enhances the final output's smoothness.}
\resizebox{\columnwidth}{!}{
\begin{tabular}{l c c c c }
\toprule
\multicolumn{1}{c}{} & \multicolumn{2}{c}{\textbf{Breath-Hold Button}} & \multicolumn{2}{c}{\textbf{Free breathing}} \\ 
\multicolumn{1}{l}{\textbf{Method}} & \textbf{no INR} & \textbf{with INR} & \textbf{no INR} & \textbf{with INR} \\ 
\midrule
\multicolumn{1}{l}{\textbf{Laplacian average V1}} & 0.627  & \textbf{0.296}  & 0.517 &  \textbf{0.251}\\ 
\midrule
\multicolumn{1}{l}{\textbf{Laplacian median V1}} & 0.452 & {\bf 0.298} & 0.396 & {\bf 0.247} \\ 
\midrule
\multicolumn{1}{l}{\textbf{Laplacian average V2}} & 0.639  & \textbf{0.432}  & 0.641 &  \textbf{0.424}\\ 
\midrule
\multicolumn{1}{l}{\textbf{Laplacian median V2}} & 0.460 & {\bf 0.430} & 0.506 & {\bf 0.416} \\ 
\bottomrule
\end{tabular}
}
\label{Tab: Results Phantom Detection}
\end{table}
In Fig.~\ref{Fig:breathing_mode}, the results from the free-breathing experiment are presented. 
The top two volumes show the 3D aorta volume without any breathing compensation and the parts from exhale phase only are shown in red. The two lowermost 3D reconstructions compare traditional linear interpolation with the result from our network-driven approach. Once again, our method yields a better reconstruction and smoothly fills in the gaps. In comparison, the linear interpolation approach struggles with larger gaps, resulting in pronounced step edges in the final reconstruction.

Table \ref{Tab: Results Phantom Detection} provides a quantitative assessment of mesh smoothness under the two breathing modes: breath-hold button and free breathing. The metrics used are the average and median magnitudes of the Laplacian vectors over the entire mesh, which serve as indicators of the overall surface roughness or smoothness. Notably, for both breathing modes, the output of the INR model has lower Laplacian roughness values for both volunteers, indicating smoother meshes. Specifically, under the breath-hold mode of the first volunteer, the average roughness decreased from 0.627 without INR to 0.296 with INR, and similarly, for the free-breathing mode, from 0.517 to 0.251. The results for the second volunteer follow a similar trend. Moreover, both the average and median roughness values are similar, which suggests a more uniform curvature distribution across the mesh. These findings highlight the efficacy of our INR pipeline in enhancing the smoothness of the reconstructed meshes, irrespective of the breathing mode.

From our evaluation, it's evident that prioritizing images from the exhale phase significantly enhances the quality of 3D reconstructions, mitigating the distortions and noise introduced by breathing movements. Our innovative INR pipeline, when compared to traditional methods, consistently showcased superior performance in both the breath-hold and free-breathing experiments. The breath-hold experiment, which closely mirrors an idle state, served as an effective benchmark, reinforcing the efficacy of our approach. Particularly in the free-breathing experiment, our method adeptly handled the challenges posed by respiratory movements, producing smoother reconstructions and effectively interpolating missing gaps. In contrast, conventional linear interpolation techniques exhibited limitations, especially in handling larger gaps, leading to pronounced step edges in the reconstructions. These findings underscore the potential of our INR pipeline in advancing the field of 3D ultrasound imaging, especially in scenarios where breathing-induced distortions are a concern. Future work aims to enhance the pipeline's adaptability to diverse anatomical variations and its compatibility with various ultrasound probes and machines, broadening its applicability in clinical settings.

\section{CONCLUSIONS}

The advancements in ultrasound imaging and its advantages over other imaging modalities have led to its wide usage in diagnostics and screening. However, challenges such as inter-operator variability and breathing-induced motion limit its applications. In this work, we introduced a novel approach leveraging INRs to effectively counteract the distortions caused by breathing when constructing a 3D volume. Furthermore, we combined this method with a robotic ultrasound system, allowing for standardized acquisitions. The screening of the abdominal aorta is targeted as a use case to verify the efficacy of the approach. The resultant smoother 3D reconstructions, as showcased by our experiments, can help clinicians visualize and inspect the progression of a disease and perform diameter measures more easily. The findings from our experiments underscore the potential of our approach to enhance clarity and improve diagnostic accuracy using 3D ultrasound images.





\vfill
\newpage
\bibliographystyle{unsrt}
\bibliography{refs}
\vfill
\newpage


\end{document}